\pdfoutput=1

\documentclass[11pt]{article}

\usepackage[final]{acl}

\usepackage{times}
\usepackage{latexsym}

\usepackage[T1]{fontenc}

\usepackage[utf8]{inputenc}

\usepackage{microtype}

\usepackage{inconsolata}

\usepackage{graphicx}
\usepackage{cite}
\usepackage{amsmath,amssymb,amsfonts}
\usepackage{algorithmic}
\usepackage{graphicx}
\usepackage{textcomp}
\usepackage{xcolor}

\usepackage{multirow}
\usepackage{adjustbox}
\usepackage{arydshln}
\usepackage{hyperref}
\def\BibTeX{{\rm B\kern-.05em{\sc i\kern-.025em b}\kern-.08em
    T\kern-.1667em\lower.7ex\hbox{E}\kern-.125emX}}

\author{Supriya Manna   {\normalfont and}  Niladri Sett\thanks{Corresponding author}\\SRM University AP, India\\\texttt{reachsmanna@gmail.com, settniladri@gmail.com}} 

\title{Faithfulness and the Notion of Adversarial Sensitivity in NLP Explanations}
\begin{document}
\maketitle

\begin{abstract}
Faithfulness is a critical metric to assess the reliability of explainable AI. In NLP, current methods for faithfulness evaluation are fraught with discrepancies and biases, often failing to capture the true reasoning of models. We introduce \textit{Adversarial Sensitivity} as a novel approach to faithfulness evaluation, focusing on the explainer's response when the model is under adversarial attack. Our method accounts for the faithfulness of explainers by capturing sensitivity to adversarial input changes. This work addresses significant limitations in existing evaluation techniques, and furthermore, quantifies faithfulness from a crucial yet underexplored paradigm.
\end{abstract}

\section{Introduction}

Deep learning-based Language Models (LMs) are increasingly used in high-stakes Natural Language Processing (NLP) tasks~\citep{minaee2021deep, samant2022framework}. However, these models are extremely opaque. To build user trust in these models' decisions, various post-hoc explanation methods~\citep{madsen2022post} have been proposed \citep{jacovi2021formalizing}. Despite their popularity, these explainers are frequently criticized for their `faithfulness', which is loosely defined as how well the explainer reflects the underlying reasoning of the model~\citep{lyu2024towards,jacovi2020towards}. In the context of NLP, explainers assign weights to each token indicating their importance in prediction, and faithfulness is measured by how consistent these assignments are with the model's reasoning. However, since the explainer is not the model itself \citep{rudin2019stop}, practitioners have developed several heuristics to measure the quality of these assignments \citep{deyoung2019eraser, zhou2022feature, nguyen2018comparing, jain2019attention, hooker2019benchmark,lyu2024towards}.

A common assumption behind many of these heuristics is the \textit{linearity} assumption, which posits that the importance of each token is independent of the others \citep{jacovi2020towards}. Based on this, a group of practitioners hypothesised that removing important tokens indicated by a faithful explainer should change the prediction, whereas removing the least important ones should not. Jacovi et al. \citep{jacovi2020towards} addressed these as \textit{erasure}. DeYoung et al. \citep{deyoung2019eraser} generalize the same with comprehensiveness and sufficiency. However, it has been exhaustively shown that the removal of features can produce counterfactual inputs\footnote{Counterfactual inputs (CI) \& counterfactual explanations (CE) are completely different. Removing features from the main input makes CI wrt the actual input. Miller et al. used this terminology \citep{miller2019explanation}. We've discussed CE in Section~\ref{s5}. Hase et al. \citep{hase2021out} debunked the same confusion of the reviewers \href{https://openreview.net/forum?id=HCrp4pdk2i&noteId=HHS46Nzr6PN}{here.} } that are out of distribution~\citep{hase2021out, chrysostomou2022empirical,lyu2024towards,janzing2020feature,haug2021baselines,chang2018explaining}, socially misaligned \citep{jacovi2021aligning}, and often severely \textit{pathogenic} \citep{feng2018pathologies}. Furthermore, evaluation metrics such as \textit{Area Under the Perturbation Curve} (AUPC) \citep{samek2016evaluating} are suspected to be severely \textit{misinformative} \citep{ju2021logic}. Instead of evaluating faithfulness, these methods primarily compute the similarity between the evaluation metric and explanation techniques, assuming the evaluation metric itself to be the ground truth \citep{ju2021logic}.

Another line of work, known as adversarial robustness~\citep{baniecki2024adversarial}, assumes that similar inputs with similar outputs should yield similar explanations. However, Ju et al.\citep{ju2021logic} has empirically shown that the change in attribution scores may be because the model's reasoning process has genuinely changed, rather than because the attribution method is unreliable. Moreover, this assumption is mainly valid when the model is `astute' \citep{bhattacharjee2020non, khan2024analyzing} and doesn't necessarily apply to explainers that don't perform local function approximation for feature importance estimation \citep{han2022explanation}. As a result, this assumption is practically \textit{restrictive} and \textit{vague}, leading practitioners to hesitate in endorsing this approach for assessing faithfulness \citep{lyu2024towards}.

Across almost all popular lines of thought, the settings in which faithfulness is quantified are \textit{linear} \citep{jacovi2020towards}, \textit{restrictive} \citep{khan2024analyzing}, \textit{misinformative} \citep{ju2021logic}, and thus the judgements on explainer quality based on such quantification could be arguable. Since, understanding the model's reasoning is challenging, and aforementioned assumptions are often deceptive, in this work, we take a fundamental approach. Previous research has demonstrated that deep models are not only opaque but also severely fragile \citep{goodfellow2014explaining,szegedy2013intriguing}. As explainers are primarily to facilitate \textit{trust} on these complex models, we argue that a faithful explainer is obligated to uncover such vulnerabilities and anomalous behaviour of the model to the end user. In this context, we introduce the notion of `adversarial sensitivity' for the explainers.
We seek the most similar (semantically and/or visually) counterpart(s) from the entire input space (subjected to certain constraints) that produces a different output, aka `adversarial examples' \citep{goodfellow2014explaining}. These pairs of inputs are always bounded by a certain distance, ensuring they are sufficiently comparable. Consequently, unlike counterfactuals, these pairs are much less likely affected by abrupt \textit{semantic shifts} \citep{lang2023surveyoutofdistributiondetectionnlp} that often lead to out-of-distribution scenarios \citep{hendrycks2016baseline, sun2022dice, sun2021react, liang2017enhancing}, making our comparisons more nuanced and robust. However, since these pairs yield different outputs, their underlying reasoning in the model is bound to differ \citep{jacovi2020towards, adebayo2018sanity}. Faithful explanations should reflect these changes, highlighting the difference in the model's inherent reasoning. We formally define the same as `adversarial sensitivity' of the explainers. Our contributions in this paper are summarised as:
\begin{itemize}
    \item we introduce the notion of `adversarial sensitivity' of an explainer, and propose a necessary test for faithfulness based on it;
    \item we present a robust experimental framework to conduct the faithfulness test;
    \item we conduct the proposed faithfulness test on six state-of-the-art post-hoc explainers over three text classification datasets, and report its (in)consistency with popular erasure based tests.
\end{itemize}

This paper is organised as follows: We introduce the notion of adversarial sensitivity, exploring its significance and relation with faithfulness in Section \ref{s2}. In Section \ref{s3}, we details our methodology, outlining the framework used to conduct our investigations. In Section \ref{s4}, we present our findings, offering in-depth analysis and interpretations of the data. We contextualize our work within the broader research landscape in Section \ref{s5}, highlighting how our study contributes to and extends existing knowledge. Finally, in Section \ref{s6}, we conclude by summarizing our key findings and proposing directions for future research, emphasizing the potential avenues for further exploration.
\section{Adversarial Sensitivity}\label{s2}
In this section, we introduce the notion of adversarial sensitivity, exploring its significance and relation with faithfulness. Thereafter, we propose the guideline for evaluating faithfulness with adversarial sensitivity.

\textbf{Definition 1.} \textit{Adversarial Example} (AE): Given a model \( f: X \rightarrow Y \), where \( X \) is the space of textual inputs and \( Y \) is the set of classes, if there exists \( x' \) for a given input \( x \in X \) such that:
\[
\{ x' \in X \mid S(x, x') \geq \theta \text{ and } f(x) \neq f(x') \},
\]
we call \( x' \) an \textit{adversarial example} (AE), where \textit{S}($\cdot$,$\cdot$) is a similarity measure and $\theta$ is a predefined similarity threshold.\newline
\textbf{Definition 2.} \textit{Local Explanation}: A local feature importance function $I$ takes an instance $x \in X$ and the model $f$ as input, and produces a weight vector as output: \[ I(f, x) = W_{x,f} = (w_1, w_2, \ldots, w_n), \]
where $w_i$ represents the \textit{importance} of the $i$-th token $x_i$ for the prediction $f(x)$.\newline
\textbf{Definition 3.} \textit{Adversarial Sensitivity}: Adversarial Sensitivity for a local explainer $\mathit{I}$ for (\(x, x'\)) is given by \(\textit{d}(W_{x,f}, W_{x',f})\). Here, \(x'\) is an AE of input \(x\), and $\textit{d}(\cdot,\cdot)$\footnote{in details at Section~\ref{s3}} is a distance measure.\newline
\textbf{Adversarial Sensitivity and Faithfulness}: Given $x'$ is an AE of \textit{x} for \textit{f}, if \textit{I} is `faithful' to \textit{f}, then $W_{x,f}$ and $ W_{x',f}$ should be dissimilar. In our setup, we report the mean distance over all obtained pairs of (\(x, x'\)).  This is a necessary but not sufficient condition for faithfulness. Currently (at the time of writing this paper) there is no necessary and sufficient condition for faithfulness \citep{lyu2024towards}. However, following the argument of Lyu et al. \citep{lyu2024towards}, as these metrics are primarily (meta)heuristic based evaluations, accessing faithfulness with several necessary tests is much more practical than attempting to formulate an exhaustive list of necessary and sufficient conditions and then evaluating against all of them. Adversarial Sensitivity is one of such necessary tests to evaluate the \textit{faithfulness} of explainers.

Adversarial Machine Learning research has extensively demonstrated that even minimal perturbations in the input space can deceive well-trained models \citep{alzantot2018generating, garg2020bae, li2020bert, gao2018black, ebrahimi2017hotflip, kuleshov2018adversarial, zang2019word, pruthi2019combating, jin2020bert, li2018textbugger, ren2019generating}. Given the discrete and combinatorially large nature of the input space, finding all possible adversarial examples (AEs) under all possible constraints is often impractical, especially in a black-box setting. Therefore, we advocate for greedily searching for AEs within a well-tested set of constraints to avoid obfuscating and low-quality examples. In this study, we select extensively used word-level, character-level, and behavioural invariance constraints. Whether methods like back-translation, paraphrasing, or hybrid attacks etc \citep{zhang2020adversarial} maintain semantic and structural similarity while generating AEs, and suitability for faithfulness evaluation are kept for further study. 

Obtaining AEs is conducted in two ways: assuming the model to be either white-box or black-box. In a white-box setting, gradients are primarily used first to identify the importance of tokens and then perturb them to create an adversarial input if the output changes. For our setting, this approach has two distinct problems. Firstly, gradient-based feature importance can be untrustworthy and manipulative \citep{wang2020gradient, feng2018pathologies}. Secondly, a class of post-hoc explainers (e.g., Gradient, Integrated Gradient) also uses the gradient to retrieve the importance of tokens. Comparing these with explainers that do not use gradient information, such as LIME or SHAP, may lead to biased comparisons. Lastly, popular gradient-based attacks such as HotFlip \citep{ebrahimi2017hotflip} are often less likely to adhere to perturbation constraints while crafting adversarial examples \citep{wang2020gradient}. Therefore, we do not consider investigation on white-box attacks for \textit{adversarial sensitivity} and adhere to a more practical, model-agnostic, and transferable black-box attacking framework. However, even in the black-box settings we employ some ad-hoc heuristics for greedily perturbing the words based on its relative importance \citep{zhang2020adversarial}, but modern explainers do not use such ad-hoc methods for calculating feature importance \citep{lyu2024towards,madsen2022post}. Therefore, our faithfulness test is unbiased towards the underlying mechanisms of (almost) all types of modern post-hoc explainers.

\section{Faithfulness Test Setup}\label{s3}
\subsection{Obtaining AEs}
\label{s3-1}
Primarily, obtaining AEs (an \textit{adversarial attack} on the model) is a greedy or brute-force procedure, where a search algorithm iteratively selects locally optimal constrained perturbations until the label changes \citep{morris2020textattack}. As mentioned in Section \ref{s2}, we devise our attacks in three constraint classes: word level, character level, and behaviorul invarince. We brief the implementation details of these attacks as follows.
\subsubsection{Word Level (\textit{A1})}
We adhere to the constraints proposed in the strong baseline `TextFooler' \citep{jin2020bert} while implementing our word-level attack (\textit{A1}). Initially, we assign weights to each word based on its impact on the model's prediction when removed. Then, in decreasing order of importance, we take each word (except stopwords), find semantically and grammatically correct $K$ (we set $K=50$) words to replace the selected word, and generate all possible intermediate corpus and query the model. If the best result (which alters the prediction the most) from this pool exceeds the one from the previous iteration, we select the new one as the current result; otherwise, we stick to the previous one. This process iterates until the current result yields a different output or we have exhaustively searched the set of possible results and found none that alter the output.\newline 
Although the constraints, including vocabulary selection and stopwords filtering, were effective in crafting adversarial examples, we observed some discrepancies with off-the-shelf hyperparameter selections. Consequently, we adjusted the minimum word embedding cosine similarity to 0.5 (instead of 0.7) and set an angular similarity threshold of 0.84 within a 15-token window.\footnote{We discovered that the authors of TextAttack \citep{morris2020textattack} identified bugs in the original implementation of TextFooler \citep{jin2020bert} and suggested a set of hyperparameters that were mostly coherent in our setup. Details can be found \href{https://textattack.readthedocs.io/en/master/_modules/textattack/attack_recipes/textfooler_jin_2019.html\#TextFoolerJin2019}{here}.}
\subsubsection{Character Level (\textit{A2})}
For character-level attack (\textit{A2}), we assign weights to each word based on its impact on the model's prediction when replaced with an unknown token (`[UNK]'). The rest of the procedure is the same as the \textit{A1}, but instead of semantically similar words, we replace the selected word after applying a combination of character-level perturbations proposed by Gao et al. \citep{gao2018black}, subject to a predefined edit distance threshold,  proposed in \citep{gao2018black}. Li et al. \citep{li2018textbugger} empirically showed that character-level perturbation can change semantic alignment in the embedding space. Therefore, after filtering with edit distance, we also employ the universal sentence encoder \citep{cer2018universal} and use the similarity threshold proposed in \citep{li2018textbugger} to select the final candidate.
\subsubsection{Behaviorul Invarience (\textit{A3})}
Recently, Ribeiro et al. \citep{ribeiro2020beyond} emphasised that models are hypersensitive not only to minute perturbations but also to `invariant' tokens. Ribeiro et al. proposed `Checklist' that   evaluates models across diverse linguistic capabilities such as vocabulary, syntax, semantics, and pragmatics. For our setting, we adopt the `Invariance Testing' they proposed (\textit{A3}). We change names, locations, numbers, etc., wherever feasible in the sentences and check if these alterations affect the prediction. As Ribeiro et al. \citep{ribeiro2020beyond} showed, a model should not be sensitive to such parameters. If it is, it indicates an inability to handle commonly used linguistic phenomena, which are subsequently characterised as a type of adversarial example \citep{morris2020textattack}. We employ the off-the-shelf implementation of the invariance testing from `TextAttack' \citep{morris2020textattack}. In our datasets, we do not have a lot of instances where phone numbers, locations, age etc are present and as we are changing these only once in this attack (else it could lead to an infinite loop), the success rate of this attack is lesser than the other attacks. However, from a linguistic perspective, this attack is crucial to make our experiments exhaustive. 
\newline
In all these attacks, we do not perturb stop-words. Next, we only consider the example as successful AE if the prediction confidence crosses a certain threshold (we set it to be at least 70\%). Finally, as we are conducting model-agnostic attacks, we acknowledge that even if the constraints are reasonably restrictive, there is always a chance that any of these examples could be out-of-distribution (OOD). To mitigate such issues, we follow a robust baseline wherever required for detecting OOD scenarios by computing the `maximum/predicted class probability' (MCP) from a softmax distribution for the predicted class of each AE \citep{hendrycks2016baseline}. MCP has been evaluated as a strong baseline, particularly when the underlying model is fine-tuned (e.g., BERT, RoBERTa) \citep{hendrycks2020pretrained, desai2020calibration}. We empirically selected only those adversarial examples that had a probability exceeding 70\% across all attacks and datasets.

\subsection{Measuring the Distance}
To measure the dissimilarity of the explanations, we follow the distance measure given by Ivankay et al. \citep{ivankay2022fooling}, that is:
\begin{equation}
    d = 1 - \frac{\tau(W_{x,f}, W_{x',f}) + 1}{2}
    \label{eq:adversarial_sensitivity}
\end{equation}
where $\tau(\cdot,\cdot)$ is a correlation measure. Ivankay et al. \citep{ivankay2022fooling} chose Pearson correlation for their distance measure. But while creating adversarial examples, a common phenomenon is obtaining unequal token vectors for $(x,x')$ due to tokenisation \citep{sinha2021perturbing}.
Correlation measures like Pearson, Kendall, and Spearman cannot handle disjoint and unequal ranked lists. Sinha et al. \citep{sinha2021perturbing} used heuristics like Location of Mass (LOM) \citep{ghorbani2019interpretation} to mitigate such issues. But Burger et al. \citep{burger2023your} highlighted their shortcomings and employed Rank Based Overlap (RBO) \citep{webber2010similarity} metric. While RBO may be robust, it introduces complications, particularly with its selection of free parameter `$p$' determining the \textit{user persistence}.\footnote{Burger et al. \citep{burger2023your} used LIME's feature importance along with explanation's average length to determine the value of `$p$' for their experimentation and Goren et al. \citep{goren2018ranking} apparently used an ad-hoc value of $p =.7$ in their experimental setup.} 
Moreover, the assumption on the \textit{depth} in RBO using Bernoulli's random variable and \textit{weights} of overlaps in explanation using geometric distribution may not be always adequate as per our setting. Furthermore, the selection between the base and extrapolated versions of RBO gives rise to the disparity in `sensitivity', especially when the \textit{residual} is significant \citep{webber2010similarity}. Following the arguments of Jacovi et al. \citep{jacovi2020towards} we, do not endorse unnecessary human intervention in faithfulness studies. As RBO inherently carries the notion of the persistence of \textit{users}, we didn't select RBO for this work.

We have extensively investigated selecting the similarity measures in previous works, but none of the works has tackled the problem of unequal and/or disjoint rank lists from an axiomatic perspective that will be adequate for our setting. 
Emond et al. \citep{emond2002new} proposed a new correlation coefficient designed to accommodate incomplete and non-strict rankings; however, this metric is not considered due to the lack of formal proof or empirical evidence. Later, Monero et al. \citep{moreno2016axiomatic} introduced essential axioms for a distance measure between incomplete rankings, establishing the existence and uniqueness of such a measure and demonstrating its superiority in generating intuitive consensus rankings compared to alternative methods. Following these axioms,  we adopt the nonparametric correlation coefficient `$\hat{\tau}_x$' presented in Yoo et al. \citep{yoo2020new}, which highlights the inadequacy of the $\tau_x$ ranking correlation coefficient devised in \citep{emond2002new} in ensuring a neutral treatment of incomplete rankings. Moreover, our employed non-parametric correlation coefficient `$\hat{\tau}_x$' is a generalization of Kendall $\tau$ on the aforementioned axiomatic foundation established by Monero et al. \citep{moreno2016axiomatic} for handling a variety of ranking inputs, including incomplete and non-strict ones. Therefore, $\hat{\tau}_x$ is foundationally much robust and can handle several types of tokenization discrepancies. Furthermore, this very distance is a nonparametric generalization of the kemeny-snell distance \citep{kemeny1962preference} for nonstrict, incomplete ranking space \citep{moreno2016axiomatic}. As a result, unlike the previous distance metric, `$\hat{\tau}_x$' is not only robust but also enjoys the properties that the Kemeny-snell distance retains for all types of rankings produced by the tokenisers.

\subsection{Interpreting the Results}
Our proposed test is a necessary test for faithfulness based on the desideratum that the explainers should produce \textit{different} explanations for AEs. Obtaining AEs is always subject to different sets of constraints. As a result, each attack type i.e. \textit{A1}, \textit{A2}, \textit{A3} is disjoint in nature thus, each of them independently conducts a necessary test given they produce \textit{successful} AEs. Theoretically, there can be finitely many AEs if we keep changing the set of constraints but in this paper, we followed three extensively evaluated, diverse sets of constraints to empirically demonstrate the adversarial sensitivity of explainers around these disjoint constraint sets. As a result, our setup consists of three disjoint necessary tests for inspecting faithfulness using adversarial sensitivity. We evaluate the explainers on the basis of how much sensitivity they obtain for how many number of discrete constraint sets. However, as these are all necessary tests, the primary objective is to reject the unfaithful ones. Also, it is highly seek-worthy that explainers perform \textit{consistently} well across constraint sets. Now, if the results across constraint sets are fluctuating for a given setup, it could be confusing for the end user to evaluate the explainers holistically. This is why, for an aggregated ranking we recommend using a consensus aggregation (e.g., Kemeny-young aggregation \citep{kemeny1959mathematics}) over empirical evaluation. Although, in our experiments, we obtained consistent results across \textit{A1, A2, A3}.
\section{Experimental Setup}
\subsection{Datasets and Models}
We conducted our experiments on SST-2 \citep{socher2013recursive}, and Tweet-Eval (Hate)~\citep{barbieri2020tweeteval} for binary classification, and on AG News~\citep{zhang2015character} for multi-class classification. We fine-tuned a Distill BERT and a BERT-based model~\citep{devlin2018bert} until it achieved a certain level of accuracy for each dataset, and attacked it with the three attack methods \textit{A1, A2, and A3} described in the Section~\ref{s3-1}. We report the models' accuracy before and after each attack\footnote{If AE is not obtained, the attack is failed and vice-versa.} in Table \ref{accuracy table}. We've addressed `Tweet-Eval (Hate)' as `Twitter' throughout the paper and Distill BERT-based model \citep{sanh2019distilbert} as DERT in Table \ref{accuracy table}. We used the standard train, test split for each dataset from the huggingface library and reported results up to the second decimal place.



\begin{table*}[ht]
\centering
\caption{Accuracy, before and after attacks -- Distill BERT and BERT}
\label{accuracy table}
\label{eq:adversarial_sensitivity_combined}
\begin{adjustbox}{max width=.9\textwidth}
\begin{tabular}{l l c c c c}
\hline
\textbf{Model} & \textbf{Dataset} & \textbf{Accuracy (\%)} & \textbf{Accuracy after \textit{A1} (\%)} & \textbf{Accuracy after \textit{A2} (\%)} & \textbf{Accuracy after \textit{A3} (\%)} \\
\hline
\multirow{3}{*}{\rotatebox{90}{\textbf{DERT}}} & \textit{SST-2}   & 91.50  & 8.42  & 21.61  & 99.62  \\
                                                        & \textit{AG News} & 93.10  & 26.71 & 68.91  & 91.46  \\
                                                        & \textit{Twitter} & 51.7& 18.84 & 8.28   & 96.93  \\
\hline
\multirow{3}{*}{\rotatebox{90}{\textbf{BERT}}}          & \textit{SST-2}   & 92.43 & 10.77 & 19.00     & 99.42  \\
                                                        & \textit{AG News} & 94.40 & 25.00    & 32.00     & 94.50  \\
                                                        & \textit{Twitter} & 54.32& 23.58& 12.61& 95.29\\
\hline
\end{tabular}
\end{adjustbox}
\end{table*}

\subsection{Explainers and Faithfulness Metrics Details}
Commonly used post-hoc local explainers can be broadly categorised in two types: perturbation-based and gradient-based explainers \citep{madsen2022post}. We have considered two commonly used perturbation-based model agnostic explainers: LIME (\textit{LIME}) \citep{ribeiro2016should} and SHAP (\textit{SHAP}) \citep{lundberg2017unified}. For \textit{SHAP}, we use the default selection of partition shap. \footnote{\href{https://shap.readthedocs.io/en/latest/generated/shap.PartitionExplainer.html\#shap.PartitionExplainer}{ partition shap documentation: }} From gradient based ones, we have chosen Gradient (\textit{Grad.}) \citep{simonyan2013deep}, Integrated Gradient (\textit{Int. Grad.}) \citep{sundararajan2017axiomatic} and their \textit{xInput} version: Gradient × Input (\textit{Grad. × Input}), and Integrated Gradient × Input (\textit{Int. Grad. × Input}). 
We compare our findings with extensively used erasure \citep{jacovi2020towards} based metrics: comprehensiveness, sufficiency \citep{deyoung2019eraser}, and correlation with `Leave-One-Out' scores \citep{jain2019attention} for faithfulness comparison.
The Appendix contains the description of erasure-based faithfulness metrics and post hoc explainers used in our experiments.

We run our experiments on an NVIDIA DGX workstation, leveraging Tesla V100 32GB GPUs. We use \textit{ferret} with default (hyper)parameter selection \citep{attanasio2022ferret} for both erasure metrics and explanation methods, \textit{TextAttack} \citep{morris2020textattack}, \textit{universal sentence encoder} \citep{cer2018universal} across attacking mechanism. We wrote all experiments in Python 3.10. Our total computational time to execute all experiments is roughly 18 hours. We report the consolidated findings for both models below in Table \ref{r}.
\begin{table*}[ht]
\centering
\caption{Consolidated Findings}
\label{r}
\begin{adjustbox}{max width=\textwidth}
\begin{tabular}{c|c|cccccc|cccccc|cccccc}
& \textbf{ } & \multicolumn{6}{c}{\textit{SST-2}} & \multicolumn{6}{c}{\textit{AG News}} & \multicolumn{6}{c}{\textit{Twitter}} \\ \hline
\textbf{Model} & \textbf{Explainer} & \multicolumn{3}{c|}{\textbf{Erasure}} & \multicolumn{3}{c|}{\textbf{Adv. Sens. $\uparrow$}} & \multicolumn{3}{c|}{\textbf{Erasure}} & \multicolumn{3}{c|}{\textbf{Adv. Sens. $\uparrow$}} & \multicolumn{3}{c|}{\textbf{Erasure}} & \multicolumn{3}{c}{\textbf{Adv. Sens. $\uparrow$}} \\ \cline{3-20}
&  & \textit{Comp. $\uparrow$} & \textit{Suff. $\downarrow$} & \textit{LOO $\uparrow$} & \textit{A1} & \textit{A2} & \textit{A3} & \textit{Comp. $\uparrow$} & \textit{Suff. $\downarrow$} & \textit{LOO $\uparrow$} & \textit{A1} & \textit{A2} & \textit{A3} & \textit{Comp. $\uparrow$} & \textit{Suff. $\downarrow$} & \textit{LOO $\uparrow$} & \textit{A1} & \textit{A2} & \textit{A3} \\ \hline

\multirow{6}{*}{\rotatebox[origin=c]{90}{\texttt{DERT}}} & {\textit{LIME}} & 0.72 & 0.02 & 0.32 & 0.77 & 0.72 & 0.81 & 0.68 & -0.03 & 0.21 & 0.66 & 0.64 & 0.72 & 0.89 & 0.00 & 0.37 & 0.77 & 0.75 & 0.83 \\
& {\textit{SHAP}} & 0.70 & 0.02 & 0.27 & 0.76 & 0.74 & 0.80 & 0.63 & -0.03 & 0.13 & 0.64 & 0.61 & 0.70 & 0.85 & 0.00 & 0.33 & 0.76 & 0.73 & 0.84 \\
& {\textit{Grad.}} & 0.37 & 0.10 & 0.10 & 0.18 & 0.2 & 0.07 & 0.44 & 0.03 & 0.06 & 0.21 & 0.23 & 0.13 & 0.76 & 0.07 & 0.13 & 0.18 & 0.2 & 0.09 \\
& {\textit{Int. Grad.}} & 0.20 & 0.32 & -0.04 & 0.56 & 0.55 & 0.55 & 0.03 & 0.27 & -0.04 & 0.52 & 0.53 & 0.54 & 0.26 & 0.50 & -0.03 & 0.56 & 0.55 & 0.58 \\
& {\textit{Grad. x Input}} & 0.17 & 0.35 & -0.12 & 0.71 & 0.63 & 0.83 & 0.04 & 0.23 & -0.11 & 0.59 & 0.57 & 0.69 & 0.29 & 0.43 & -0.10 & 0.71 & 0.67 & 0.82 \\
& {\textit{Int. Grad. x Input}} & 0.53 & 0.08 & 0.24 & 0.76 & 0.70 & 0.80 & 0.54 & 0.00 & 0.12 & 0.58 & 0.57 & 0.54 & 0.81 & 0.02 & 0.22 & 0.76 & 0.72 & 0.84 \\ \hline

\multirow{6}{*}{\rotatebox[origin=c]{90}{\texttt{BERT}}} & {\textit{LIME}} & 0.68 & 0.01 & 0.33 & 0.74 & 0.75 & 0.86 & 0.72 & -0.06 & 0.14 & 0.64 & 0.54 & 0.68 & 0.86 & 0.00 & 0.32 & 0.76 & 0.75 & 0.82 \\
& {\textit{SHAP}} & 0.61 & 0.02 & 0.26 & 0.71 & 0.70 & 0.84 & 0.67 & -0.05 & 0.11 & 0.62 & 0.52 & 0.67 & 0.87 & 0.01 & 0.35 & 0.80 & 0.79 & 0.86 \\
& {\textit{Grad.}} & 0.36 & 0.09 & 0.10 & 0.17 & 0.18 & 0.04 & 0.51 & 0.04 & 0.03 & 0.23 & 0.34 & 0.14 & 0.78 & 0.05 & 0.16 & 0.16 & 0.17 & 0.07 \\
& {\textit{Int. Grad.}} & 0.19 & 0.29 & 0.00 & 0.53 & 0.55& 0.52 & 0.04 & 0.26 & -0.03 & 0.51 & 0.50 & 0.50 & 0.22 & 0.38 & -0.04 & 0.51 & 0.52 & 0.51 \\
& {\textit{Grad. x Input}} & 0.22 & 0.27 & 0.01 & 0.66 & 0.67& 0.86 & 0.46 & 0.06 & 0.16 & 0.62 & 0.54 & 0.67 & 0.21 & 0.41 & 0.00 & 0.73 & 0.71 & 0.71 \\
& {\textit{Int. Grad. x Input}} & 0.54 & 0.06 & 0.02 & 0.76 & 0.76 & 0.85 & 0.47 & 0.04 & 0.05 & 0.56 & 0.52 & 0.56 & 0.83 & 0.01 & 0.18 & 0.75 & 0.75 & 0.74 \\ \hline
\end{tabular}
\end{adjustbox}
\end{table*}

\section{Results \& Discussion}\label{s4}
From Table \ref{r}, it is clearly observable that as per Adv. Sens., LIME, SHAP, Gradient × Input, and Integrated Gradient × Input all perform competitively across various datasets and attacks. However, the vanilla versions of gradient-based methods are not as effective. Notably, the Gradient itself exhibits the least sensitivity to adversarial inputs, followed by Integrated Gradient. Furthermore, Integrated Gradient's adv. sens. remains almost invariant to the type of attacks across all datasets, unlike comp. and suff. Interestingly, all explainers except Gradient show a drop in sensitivity in the AG News dataset across all attacks. Gradient performs best on all attacks in AG News amongst datasets. Perturbation-based explainers like LIME and SHAP are among the best performers across datasets. Gradient × Input and Integrated Gradient × Input perform well within the group of white-box explainers, with LIME and SHAP.

Under erasure methods across all datasets, Gradient is a moderately well-performing explainer, whereas Gradient × Input performs much worse. However, according to Adversarial Sensitivity, Gradient × Input is one of the best performers, with Gradient being the worst among all. Like Gradient × Input, Integrated Gradient also largely performs worse than Gradient in erasure, but it remains consistently moderate according to Adv. Sens. Both LIME and SHAP not only perform very well in both Adv. Sens. and erasure metrics but also the difference b/w their magnitudes for both erasure metrics and adv. sens. are (considerably) nominal. Integrated Gradient × Input is substantially similar to LIME, SHAP in adv sens., but we observe a considerable drop in comprehensiveness for SST-2 and AG News for both the models, unlike adv. sens. 

To demonstrate, how to evaluate the explainers based on the consensus ranking, we are considering the case of SST-2 for the BERT Model. We use the Kemeny-Young method here; as this has been extensively used for Condorcet ranking \citep{young1988condorcet}; it also satisfies highly desirable social choice properties for \textit{fair} voting \citep{owen1986information, young1995optimal}. Kemeny-Young aggregation also have been used in biology and social science extensively \citep{brancotte2014conqur,andrieu2021efficient, arrow2010handbook}. We first convert the columns of \textit{A1, A2, A3} into ranking vectors using a ranking function. In our case, we used the traditional ranking: the higher the score (here the \textit{score} is average distance obtained), the lower the ranking. We obtained the consensus ranking vector as $[2,3,6,5,4,1]$. Here, the indices of the vector denote the respective position of explainers (starting from 1 onwards) in the `\textbf{Explainer}' column.

DeYoung et al. \citep{deyoung2019eraser} advocated for both \textbf{high} comprehensiveness and \textbf{low} sufficiency for adequate explanations but unlike us; they did not propose any consensus evaluation for explainers with these two parameters taken together. According to the definition, both metrics measure two different aspects of explanations. This makes the evaluation of explainers even confusing with comprehensiveness-sufficiency, especially if the results for these two metrics are fluctuating. We did not find any axiomatically valid evaluation strategy for explainers in the presence of different kinds of faithfulness metrics in subsequent literature (including DeYoung's paper \citep{deyoung2019eraser}) as well. It is worth noting Javoci et al. \citep{jacovi2020towards} reported the same observation previously. As Javoci et al. said, "\textit{Lacking a standard definition, different works evaluate their methods by introducing tests to measure properties that they believe good interpretations should satisfy. Some of these tests measure aspects of faithfulness. These ad-hoc definitions are often unique to each paper and inconsistent with each other, making it hard to find commonalities.}" \citep{jacovi2020towards}. 

Although evaluation metrics are inherently different from one another, for the sake of demonstrating an inter-comparison between erasures and adv. sens.\footnote{we do not necessarily endorse this rank-based comparison as an axiomatic comparison in the presence of different type of faithfulness evaluation parameters but a (hard) estimate in the absence of such comparisons.}, we rank the explainers based on the \textit{scores} they obtain in individual erasure methods in the case of SST-2 for the BERT Model in Table \ref{r}. We consider the same ranking function used for adv. sens. for Comprehensiveness and LOO score and the inverse of the same ranking function for Sufficiency due to its opposite nature with respect to the former. First, we take the Kemeny-Young aggregation of comprehensiveness and sufficiency; the ranking obtained is $[1,2,4,6,5,3]$. LOO's ranking is: $[1,2,3,6,5,4]$. Next, we combine all erasure columns and get the aggregation as $[1,2,4,6,5,3]$. The obtained aggregated ranking for adv. sens. was $[2, 3, 6, 5, 4, 1]$. From this comparison, we retrieve all explainers have obtained different rankings for comprehensiveness-sufficiency, LOO, and combined aggregation of erasures, as compared with adv sens. Throughout our experiments for both models, we observed explainers except for LIME \& SHAP (as mentioned earlier) are largely inconsistent with one or more erasure method(s).

Nevertheless, erasure has been used in several novel affairs and benchmarkings \citep{mathew2021hatexplain,atanasova2023faithfulness,liu2022rethinking,babiker2023intermediate} due to its easy-to-implement and seemingly reasonable assumption. However, we observe in our experimentation that erasure methods are inconsistent except perturbation based explainers with our proposed metric. Unlike erasure, which makes simplistic assumptions about the independence of the token's importance and absence of non-sensical OOD results while removing tokens \citep{lyu2024towards}, adversarial sensitivity is founded on the assumption that faithful explainers should capture the intrinsic dissimilarity of model reasoning when \textit{fooled}. We, therefore, advocate for the adoption of adversarial sensitivity as a foundational metric for a necessary test of faithfulness for assessing explainers.


\section{Related Works}\label{s5}
\textbf{Faithfulness evaluation}, based on previous literature, can be broadly categorised in six ways: axiomatic evaluation, predictive power evaluation, robustness evaluation, perturbation-based evaluation, white-box evaluation, and human perception evaluation \citep{lyu2024towards}. The commonly used erasure is primarily a perturbation-based evaluation: it hypothesised that the change in model's output caused by the removal tokens is proportional to the \textit{importance} of the tokens for the prediction. If a local explainer is \textit{faithful}, 
removal of \textit{important} tokens as identified by the explainer should align with the hypothesis. \textit{Comp., Suff., LOO} are different instances of the erasure hypothesis. Our hypothesis is also somewhat related to the perturbation-based evaluation. We hypothesised that a \textit{ faithful} explainer should be sensitive to anomalous input that \textit{fools} the model. We perturb at several levels in the input to deceive the model, not to interpret. Next, we evaluate how much the explainer is \textit{sensitive} towards the subtle changes that deceive the model. As the deep models are known to be severely fragile, we argue this is a necessary quality for the explainer to be \textit{faithful} when the model is not showing its \textit{expected} behaviour.

Following the hypothesis of adversarial robustness, which comes under the robustness evaluation category, successful \textit{adversarial attack} on explainer aims to perturb the input such that an explainer generates dissimilar (non-robust) explanations subject to `similar' input and `similar' (bounded by a \textit{certain} distance) output \citep{baniecki2024adversarial} (AdvxAI). However, rather than any ad-hoc distance to compare the similarity of explainer Alvarez et al. \citep{alvarez2018robustness} emphasize on the (local) lipschitz continuity measurement in this setting. Khan et al. \citep{khan2024analyzing} has recently analysed the theoretical bounds of (dis)similarity under this setting when the explainer and classifier \citep{bhattacharjee2020non} are astute. Anyways,  AdvxAI is not a formally accepted measure of faithfulness \citep{ju2021logic,zhou2022exsum, lyu2024towards}, as the model may yield different reasoning rather than the explanation is non-robust. Anyhow, in this work we conduct attacks to deceive the model, not the explainer following the aforesaid hypothesis. For a broad overview on faithfulness evaluation we suggest the reader to refer to \citep{lyu2024towards}.

\textbf{Adversarial examples} (AdvAI) can be crafted at several levels: word level, character level, phrase level, paraphrasing, back translation, invariance testing, etc. in white-box and black-box settings primarily \citep{zhang2020adversarial}. We employed word level, char level and invarience testing attacks. Noppel et al.\citep{noppel2023sok} systematised the underlying relations of AdvAI and AdvxAI. For a broder overview, we refer the reader to \citep{qiu2022adversarial,zhang2020adversarial}. 
Adversarial attacks on NLP systems have been carried out primarily in 2 types: white box and black box \citep{zhang2020adversarial}, we didn't go with white box ones as they primarily leverage gradient information also, as several explainers such as Integrated Gradient or Gradient access the same information which could constitute a biased evaluation \citep{ju2021logic} as the attacking mechanism and the explanation method are similar and both leverage gradient information.

\textbf{Counterfactual explanations} \citep{mothilal2020explaining}, which demonstrate the changes would produce a distinct outcome, differ fundamentally from adversarial examples \citep{freiesleben2022intriguing}, which aim to deceive models with minimal input changes. Counterfactuals should be semantically and/or visually different \citep{yang2020generating}. Thus, it is not intended to deceive the underlying model.  In the context of Natural Language Inference (NLI), Atanasova et al. \citep{atanasova2023faithfulness} experimented with counterfactuals to investigate faithfulness. Camburu et al. \citep{camburu2019make} explored inconsistencies in explanations for NLI but did not adhere to the constraints necessary for generating adversarial inputs required in our setting. Moreover, counterfactual explanations can potentially highlight necessary features but may miss sufficient ones for prediction \citep{hsieh2020evaluations}.

\textbf{Similarity measures} in previous works, especially in AdvxAI \citep{sinha2021perturbing,burger2023your,ivankay2022fooling}, have used mainly correlations, distance measures, and top `k'\% intersection in tokens. Burger et al. \citep{burger2023your} comprehended the common issues with such metrics due to tokenization discrepancies and employed RBO \citep{webber2010similarity}. We did not select RBO having the free parameter \textit{user persistence (p)}, as we argue that faithfulness should not be based on the \textit{unnecessary} human evaluations. We rather select the distance invented by Moreno et al. \citep{moreno2016axiomatic} that satisfies all the axioms for non-strict, incomplete rankings and also satisfies the desirable social choice properties of the Kemeny-snell distance \citep{kemeny1962preference} for \textit{fair and conclusive} rankings.

\section{Conclusion and Further Work}\label{s6}
In this work, we explored the shortcomings of widely used faithfulness measures in NLP and proposed a test to evaluate explainers based on their sensitivity to adversarial inputs. Through extensive experiments on six post-hoc explainers, we found that gradient \& integrated gradient aren't (sufficiently) sensitive, while LIME, SHAP, and Gradient × Input, and Integrated Gradient × Input show better sensitivity. We also observed notable differences between our evaluation and traditional erasure-based faithfulness measures.

Future work will explore adversarial sensitivity for multilingual datasets, low-resource languages, and advanced lms.

\section*{Broader Impact}
Deep models are not only fragile but also opaque. Our work lies at the intersection of these two critical aspects. Building on the arguments presented by Jacovi et al. \citep{jacovi2020towards}, we introduce a necessary test for assessing faithfulness. Given that the underlying assumption of adversarial sensitivity is applicable to (nearly) all data types and models, this concept can be extended across (almost) all domains and explanation mechanisms.

Faithfulness is a key component in explainable AI \citep{miller2019explanation}. When a model behaves deceptively under any form of adversarial intervention, it becomes imperative that explainers provide \textit{faithful} explanations in such scenarios, rather than merely those where the model performs according to user expectations. Adversarial sensitivity aids end-users in identifying explainers that are \textit{responsive} to adversarial instances. We strongly believe that the nuanced notion of adversarial sensitivity opens up a new direction for evaluating explainers, particularly in situations where being \textit{unfaithful} could lead to a misinterpretation of why the model produces deceptive results.

\section*{Limitation}
Adversarial attacks are computationally expensive. Our work therefore is much computationally expensive and non-trivial than erasures. Our work is a necessary test faithfulness of explainers therefore, from a practitioner's perspective \citep{lyu2024towards} we employ our tests primarily to identify unfaithful explainers. It's important to note that our test does not take into account other criteria, such as biases in models, during the evaluation process. The scope of the work, for the time being, is restricted to NLP.

\section*{Acknowledgement}
The authors are thankful to Dr. Adolfo Escobedo, the co-author of \citep{yoo2020new} for sharing the code for their proposed metric ($\hat{\tau}_x$). The authors also thank the reviewers for their insightful comments.

\bibliography{latex/custom}
\clearpage
\section{Appendix}\label{app}
\subsection{Short Description of the Erasure Methods}
We compare our findings with extensively used erasure \citep{jacovi2020towards} based metrics: comprehensiveness, sufficiency \citep{deyoung2019eraser}, and correlation with `Leave-One-Out' scores \citep{jain2019attention} for faithfulness comparison.
Below are the definitions of these metrics.
\paragraph{Comprehensiveness ($\uparrow$)} This metric evaluates the extent to which an explanation captures the tokens crucial for the model's prediction. It is quantified by:

\begin{equation}
    \text{Comprehensiveness} = f_j(x) - f_j(x \setminus r_j)
\end{equation}

where $x$ is the input sentence, $f_j(x)$ is the model's prediction probability for class $j$, and $r_j$ is the set of tokens supporting this prediction. $x \setminus r_j$ denotes $x$ with $r_j$ tokens removed. A higher value indicates greater relevance of $r_j$ tokens.

For continuous feature attribution methods, we compute comprehensiveness multiple times, considering the top $k\%$ (from 10\% to 100\%, in 10\% increments) of positively contributing tokens. The final score is the average across these computations.

\paragraph{Sufficiency ($\downarrow$)} This metric assesses whether the explanation tokens suffice for the model's prediction:

\begin{equation}
    \text{Sufficiency} = f_j(x) - f_j(r_j)
\end{equation}

A lower score suggests that $r_j$ tokens drive the prediction. As in comprehensiveness, we calculate the aggregate sufficiency.

\paragraph{Correlation with Leave-One-Out scores ($\uparrow$)} We compute Leave-One-Out (LOO) scores by iteratively omitting each token and measuring the change in model prediction. LOO scores represent individual feature importance under the \textit{linearity assumption} \citep{jacovi2020towards}. We then calculate the Kendall rank correlation coefficient $\tau$ between the explanation and LOO score:

\begin{equation}
    \tau_{\text{loo}} = \text{corr}_{\text{Kendall}}(\text{explanation}, \text{LOO scores})
\end{equation}

A $\tau_{\text{loo}}$ closer to 1 indicates higher faithfulness to LOO importance. We have addressed $\tau_{\text{loo}}$ as \textit{LOO} in Table \ref{r}.
\subsection{Short Description of the Explainers}
Local Interpretable Model-agnostic Explanations (LIME), introduced by Ribeiro et al. (2016) \citep{ribeiro2016should}, operates on the principle of local approximation. LIME generates explanations by fitting interpretable models to local regions around specific instances, providing insights into the model's behavior for individual predictions. This approach is particularly valuable for understanding non-linear models in a localized context.

SHapley Additive exPlanations (SHAP), developed by Lundberg and Lee (2017) \citep{lundberg2017unified}, draws from cooperative game theory, specifically Shapley values \citep{shapley1951notes}. SHAP assigns each feature an importance value for a particular prediction, ensuring a fair distribution of the model output among the input features. This method offers a unified framework that encompasses several existing feature attribution methods.

Gradient-based attribution methods leverage the model's gradients with respect to input features to quantify their importance. The simple Gradient method \citep{simonyan2013deep} computes the partial derivatives of the output with respect to each input feature, providing a first-order approximation of feature importance. However, this approach can suffer from saturation issues in deep networks.

To address these limitations, Sundararajan et al. (2017) \citep{sundararajan2017axiomatic} proposed Integrated Gradients, which considers the integral of gradients along a straight path from a baseline to the input. This method satisfies desirable axioms such as sensitivity and implementation invariance, making it a robust choice for attribution.

Variants of these methods, namely Gradient × Input and Integrated Gradient × Input, incorporate element-wise multiplication with the input to account for feature magnitude. These approaches can provide more intuitive explanations, especially in scenarios where the input scale is significant.
\end{document}